\documentclass[man, floatsintext]{apa7}

\usepackage{placeins}
\usepackage{lipsum}
\usepackage[figuresright]{rotating} 
\usepackage[american]{babel}
\usepackage{amsmath}
\usepackage{amssymb}
\usepackage{subcaption}
\usepackage{graphicx}
\usepackage{bm}
\usepackage{algorithm}
\usepackage{bbm}
\usepackage{algpseudocode}

\usepackage{csquotes}
\usepackage[style=apa,sortcites=true,sorting=nyt,backend=biber]{biblatex}
\usepackage{dsfont}
\usepackage{bbm}
\usepackage[textwidth=2cm]{todonotes}
\DeclareLanguageMapping{american}{american-apa}

\title{ Active Feature Acquisition for Cost-Efficient Temporal Prediction with Reduced Participant Burden}
\shorttitle{Active Feature Acquisition for Temporal Prediction}

\abstract{

    Accurate forecasting of pathological outcomes is a central problem in psychology. To do so, psychologists often collect intensive longitudinal data. However, in such studies, the desire to acquire a large number of variables for the sake of accurate prediction is often counteracted by the need to minimize participant burden. Acquiring more variables per occasion can yield better predictions, but having too many acquisitions increase the risk of non-response and attrition. Longitudinal Active Feature Acquisition (LAFA) is a principled approach to resolve this conundrum. Instead of requiring responses to every item at every acquisition occasion, LAFA produces a policy that seeks to optimally select dynamic subsets of items to be acquired at each timepoint while preserving our ability to forecast a specific outcome. However, existing LAFA methods are mostly based on Neural Networks (NN) that are difficult to interpret in practice. In this work, we introduce a tree distillation method for learning an interpretable policy from NN-based LAFA networks. We validated our method through both a simulation and an empirical EMA dataset on forecasting daily alcohol consumption. In both cases, we find that we can meaningfully reduce the number of items acquired at each occasion with minimal loss in accuracy.

    \vspace{2em}

}

\authorsnames[*1,{*2,3},4,,5,5,2,1]{Yunni Qu, Bing Cai Kok, Whitney Ringwald, Grant King, Aidan Wright, Kathleen Gates, Junier Oliva}
\authorsaffiliations{{Department of Computer Science, University of North Carolina at Chapel Hill}, {Department of Psychology and Neuroscience, University of North Carolina at Chapel Hill}, {School of Social Sciences, Nanyang Technological University, Singapore}, {Department of Psychology, University of Minnesota Twin Cities}, {Department of Psychology, University of Michigan}}
\authornote{
  \textbf{*} Yunni Qu and Bing Cai Kok contributed equally to this work.}

\begin{document}
\maketitle

% To ask Yunni: what are ur planned analyses for the empirical data
% To ask Yunni: references for model distillation 
% To ask Yunni: Git repo

Technological advances have enabled psychologists to collect intensive longitudinal data at unprecedented temporal scales. For example, researchers have begun using smartphones to probe emotional processes on a daily basis across multiple weeks \parencite{wright2015daily}, track alcohol consumption in real time \parencite{ringwald2025common} and to also characterize temporal patterns of tobacco consumption \parencite{lydon2021temporal}. However, this newfound ability to acquire multiple repeated measurements at such scale comes with a practical challenge: when collecting intensive longitudinal data, one has to balance the desire to make the most accurate prediction, which aims at collecting the maximum number of variables per occasion, with the need to minimize participant burden. In most psychological studies, there is obvious value in acquiring as many variables per measurement occasion as possible. Larger sets of variables allow for deeper characterizations of each individual's psyche, and having access to a wider range of measures is conducive for exploratory research. Despite these benefits, collecting an excessively large number of measures per occasion is usually neither feasible nor recommended. Natural limits on human attention curtail the number of items that a human respondent can reasonably respond to in a given session, and this can be further exacerbated when conducting EMA studies on specialized populations, such as among children with ADHD \parencite{rosen2013know}. Moreover, respondents 
often suffer from fatigue and boredom when presented with the same measurement items repeatedly over time, with corresponding drops in the quality of responses collected \parencite{blacutt2026balancing}. 
In other words, naively maximizing the number of variables collected can ironically produce worse measures of the psychological processes of interest. 

This inherent tension between the quantity and quality of human measurement lies at the heart of Ecological Momentary Assessments (EMA). Extant practice in the field has tackled this issue mostly by curating a subset of possible measures based on domain expertise ahead of time, and then acquiring only those variables at each given measurement occasion \parencite{schneider2024just}. Such an approach can be thought of as a {\it static} variable acquisition scheme. That is, we acquire the exact same variables per person at every occasion. In this work, we seek to improve upon such static regimes when predicting a specific outcome is the focus. Such a criterion variable is often implicitly available in many EMA studies. For example, EMA studies on alcoholism frequently measure the amount of alcohol consumed per night as a target variable of interest \parencite{motschman2022alcohol}. Similarly, studies on suicidality often track suicidal ideation on top of other items measuring affect \parencite{czyz2023ecological}. When such a criterion is available, an alternative is to learn, from existing data, an {\it active} variable acquisition scheme where the pattern of variables obtained is personalized to each individual. At each measurement period, LAFA dynamically adjusts the most worthwhile measurements to be made based on the history of observed values for each individual. This is referred to in the Computer Science literature as Longitudinal Active Feature Acquisition (LAFA). 

% we propose to develop an algorithm that can learn, from existing data, an {\it active} variable acquisition scheme that is personalized to each individual. We will term our method as Longitudinal Active Feature Acquisition (LAFA). 

LAFA approaches differ from the static regime in two important ways. Firstly, these active feature acquisition algorithms can utilize demographic variables available at baseline (typically available in EMA studies during an onboarding process) to produce personalized acquisition policies. That is, across \textit{people} the pattern of variables may differ. Secondly, these policies are dynamic across \textit{time}: instead of requiring responses to every item at every measurement occasion, LAFA methods optimally select, based on previously observed values, dynamic subsets of items to be presented at each timepoint while preserving our ability to forecast a specific outcome. Importantly, such methods also differ from Computerized Adaptive Testing (CAT) that is popular in the psychometric literature \parencite{van2000computerized, schneider2024just}. Although both LAFA and CAT seek to reduce participant burden by presenting select items at each measurement occasion, they differ in that CAT is concerned primarily with the measurement precision of some latent trait(s) whereas LAFA is not. CAT procedures are often built upon an Item Response Theory (IRT) model, from which quantities such as measurement precision of the latent trait(s) are well-defined \parencite{hambleton1993comparison}. In contrast, LAFA eschews explicit modeling of latent traits and instead focuses on retaining predictive accuracy of a criterion variable while reducing participant burden. This shift in focus from measurement to prediction is mostly inspired by the design of some EMA studies where the ability to forecast an observable criterion, rather than the measurement of latent constructs, is of utmost interest. 

% Outside of psychology, the problem posed by LAFA has also gained recent traction in the machine learning literature. 

% Existing methods in that space have mostly framed the longitudinal acquisition problem as that of

The problem posed by LAFA can be formalized as a Markov Decision Process (MDP) \parencite{puterman2014markov, lamar2018markov, qin2023risk, kossen2022active}. A MDP is a framework designed to find an optimal policy given an appropriate specification of certain `state' and `action' spaces. This is particularly germane to our setup --- our goal is to obtain a policy mapping the history of variable acquisitions (i.e. `state') to decide on which variables to acquire next (i.e. `action'). This makes MDPs a natural approach to model the longitudinal acquisition process. Despite this natural framing, the computation of the MDP solutions is notoriously challenging when the state and action spaces are large \parencite{li2021active}. This is particularly so in our context: the size of the state space grows with the number of measurements and the action space is combinatorial in nature \parencite{qu2026relaxed, li2021active}. These constraints render a direct MDP modeling approach relatively ineffective for the EMA setup. To get around this, we build off previous work by considering a recent neural network approximation to the MDP problem whose solution can be obtained through standard gradient-based optimization \parencite{qu2026relaxed}. These NN-based approximations thus allow us to solve the large MDP problems that we encounter in EMA studies.

However, LAFA policies built from neural networks can suffer from a lack of interpretability. As a simple example, consider a neural network policy that takes as input a vector of demographic variables and returns a binary mask over the available EMA variables at the first measurement occasion (i.e. a value of `1' indicates that we should collect that variable and `0' indicates otherwise). For any given participant, one can feed their demographics into such a network to obtain an acquisition mask. However, such a mapping is opaque: it is not intuitively clear how the generated mask is related to the demographic variables. This opacity is only further compounded when trying to understand the acquisitions at time $t > 1$ where we take into account not just the demographic variables, but also the entire history of acquisitions. This lack of interpretability can be particularly problematic when one has to explain the acquisition process to practitioners, or when there is a need to modify the policy to satisfy legal or ethical considerations. In short, the black-box nature of NN-based LAFA methods can curtail its adoption and use in practice.

%  Furthermore, we demonstrate in this work that such a neural approximation can also be made interpretable via a process called `model distillation' (CITE). The key idea is that, barring some loss of precision, the black-box neural network policy can be approximated by interpretable models (e.g. decision trees). That is, compared to previous work, our proposed methodology marries the computational expedience of neural network policies with the interpretability afforded through tree-based model distillation.

What is needed, then, is a LAFA policy that is interpretable by construction: one that a researcher can read, defend to a review board, and hand to a research assistant as a protocol document. This requirement sits awkwardly with the computational reality described above. The tractability of NN-based LAFA methods is exactly what makes them opaque, and abandoning the neural approximation returns us to an MDP whose state and action spaces are intractable in the EMA setting. Our approach is therefore to keep the neural machinery for \emph{learning} the policy and to remove it at \emph{deployment} through `model distillation' \parencite{frosst2017distilling}, in which a black-box model (the `teacher') is used to train a transparent model (the `student') that reproduces its behavior.

% Given these challenges, we seek to develop a neural approximation to the LAFA problem that can be made interpretable via a process called `model distillation' \parencite{frosst2017distilling}. In particular, we will focus our attention on a particular type of neural network approximation called REACT \parencite{qu2026relaxed}, which stands for `Relaxed Efficient Acquisition of Context and Temporal Features'. The REACT framework is unique in that, unlike other existing LAFA methods, it can utilize demographic variables collected during onboarding (i.e. the `context' features) to produce personalized acquisition policies. Specifically, REACT learns a set of context features that informs prediction and   longitudinal feature acquisition. In this work, we develop a tree-based distillation method by building upon the REACT framework. This distillation process can be thought of as having the black-box neural policy from REACT `teaching' an interpretable tree policy (i.e. the `student'). We will call this extended interpretable framework as Tree-REACT (TREACT). The central idea is that, barring some loss of fidelity, we can approximate the neural acquisition policy with decision trees representing both the onboarding context acquisition and longitudinal acquisition phase, and the resulting policy can then be read off directly. That is, compared to previous work, our proposed methodology marries the computational expedience of neural network policies with the interpretability afforded through tree-based model distillation. 
The teacher we adopt is REACT \parencite{qu2026relaxed}, which is short for "Relaxed Efficient Acquisition of Context and Temporal Features". REACT is well suited to the EMA setting because, unlike other existing LAFA methods that only utilize longitudinal data, it uses an additional set of onboarding variables that are not time-varying to produce personalized acquisition policies: it learns a set of context features that informs both prediction and longitudinal feature acquisition. In this paper, we build upon REACT and distill a interpretable student policy, which consist of sequence of decision trees, one per measurement occasion. We call "\textbf{TRE}e \textbf{A}cquisition of \textbf{C}ontext and \textbf{T}emporal Features" (TREACT). Because REACT emits only a small number of distinct acquisition patterns in practice, each tree faces a manageable classification problem, and its leaves can be read off directly as the panel of items to administer.
The result is a policy that marries the computational expedience of neural network training with the interpretability of a tree-based protocol. Barring some loss of fidelity, which we quantify below, the acquisition decisions of the neural policy can be reproduced by a shallow tree that a domain expert can audit, amend, and execute by hand. Our contributions are as follow:

% \textcolor{red}{@Bing  Please fill in the contributions on the EMA side}
% Junier's first contribution rephrase: We formulate longitudinal active feature acquisition as an approach to adaptive psychological assessment, using information accumulated across measurement occasions to guide what to assess next. We develop this perspective for ecological momentary assessment (EMA), showing how the framework addresses methodological questions central to psychological research, including how to balance respondent burden with predictive accuracy and adapt assessments to individuals’ evolving responses.
\begin{itemize}
    \item We describe the EMA variable selection problem in the LAFA framework, whose solution can be made tractable through neural networks (e.g. via REACT).
    % \todo{rephrase; dynamic test time variable selection}
    \item We develop TREACT, a novel method for distilling a black-box longitudinal feature-acquisition policy into a sequence of decision trees, one per time point, that a researcher can read, audit, and administer without the neural model at deployment. 
    % \item TREACT fits the sequence on-policy: each tree is trained on the state distribution its predecessors actually induce, rather than on the teacher's trajectories, so the protocol is consistent with itself when executed forward.
    \item We show empirically that we can recover most of the teacher's cost/performance frontier with interpretable trees with modest depth. 
    % small enough to print.
\end{itemize}

The remainder of the paper is structured as follows. We first provide  background on the LAFA problem in the context of EMA research. In particular, we will introduce REACT \parencite{qu2026relaxed}, a state-of-the-art  NN-based LAFA method. This is followed by a description of the technical details of our method TREACT, a tree policy distillation method for LAFA building on REACT. We then report the results of a simulation study that is designed to verify that our dynamic acquisition method can recover variables of interest. An empirical application on forecasting alcohol consumption from EMA data is presented next. To close, we discuss the relation between our method and other existing approaches for dynamic feature acquisition in the psychometric literature, and highlight how our model distillation approach can provide actionable insights into the inner workings of a black-box variable selection algorithm. 

% TESTING!!! 
\begin{table}[htbp]
\centering

    \begin{tabular}{p{4cm} p{11cm}}
        \toprule
        \textbf{Terminology}  & \textbf{Description} \\ 
        \midrule
        
        Policy &  A protocol that states what should be done given the existing information (e.g. if A happens, then do B; otherwise, do C).  \\ \addlinespace
        
        Acquisition &  Collection of data for certain select variables (e.g. at $t=3$, ask the participant about their sleep and mood only). \\ \addlinespace
        
        % Planner & A model that parameterizes a policy. A simple example would be a logistic regression planner with the previous measurements as input; the logistic regression model can then output the probability of selecting a variable to be acquired. \\ \addlinespace
        
        Distillation & The process of using a simpler (and usually more interpretable) model to approximate the behavior of another model. \\  \addlinespace

        Teacher &  The teacher is the original, complex model whose decisions we want a simpler model to imitate. \\  \addlinespace

        Student &  The student is the simpler, more interpretable model trained to reproduce the teacher’s decisions. \\  
        \addlinespace

        Rollout & A rollout is the process of applying an acquisition policy sequentially to a participant, producing a series of states that contain the information acquired up to each occasion.\\
        \addlinespace

        LAFA &  Longitudinal Active Feature Acquisition; general term for method trained for dynamic test-time selction of variables. \\  
        \addlinespace
        \bottomrule
    \end{tabular}

    % Caption goes smoothly at the bottom here
    \vspace{1em}
    \caption{Glossary of terms used to describe the EMA problem in the TREACT framework.}
    \label{tab:glossary}
\end{table}

\section{REACT: Neural Network based LAFA Policy}

Consider a dataset $\mathcal{D} = \{ ( \bm{s}^{(i)}, \bm{X}^{(i)}, \bm{y}^{(i)} ) \}_{i=1}^N$ that is collected from an EMA study. Here, $\bm{s}^{(i)} \in \mathbb{R}^{d_s}$ is the vector of demographic variables for participant $i$ obtained at onboarding. $\bm{X}^{(i)} \in \mathbb{R}^{d_x \times T}$ is a matrix containing the temporal features across $T$ discrete timesteps for participant $i$. Similarly, $\bm{y}^{(i)} \in \mathbb{R}^{T}$ is the vector containing the outcome variable across the $T$ timesteps for participant $i$. For simplicity, we will assume that the outcome is discrete (i.e. $\bm{y}^{(i)}_t \in \{1, ..., C\}$). Similar ideas extend to the case where the outcome is a real number. 

\begin{figure}
    \centering
    \includegraphics[width=0.8\linewidth]{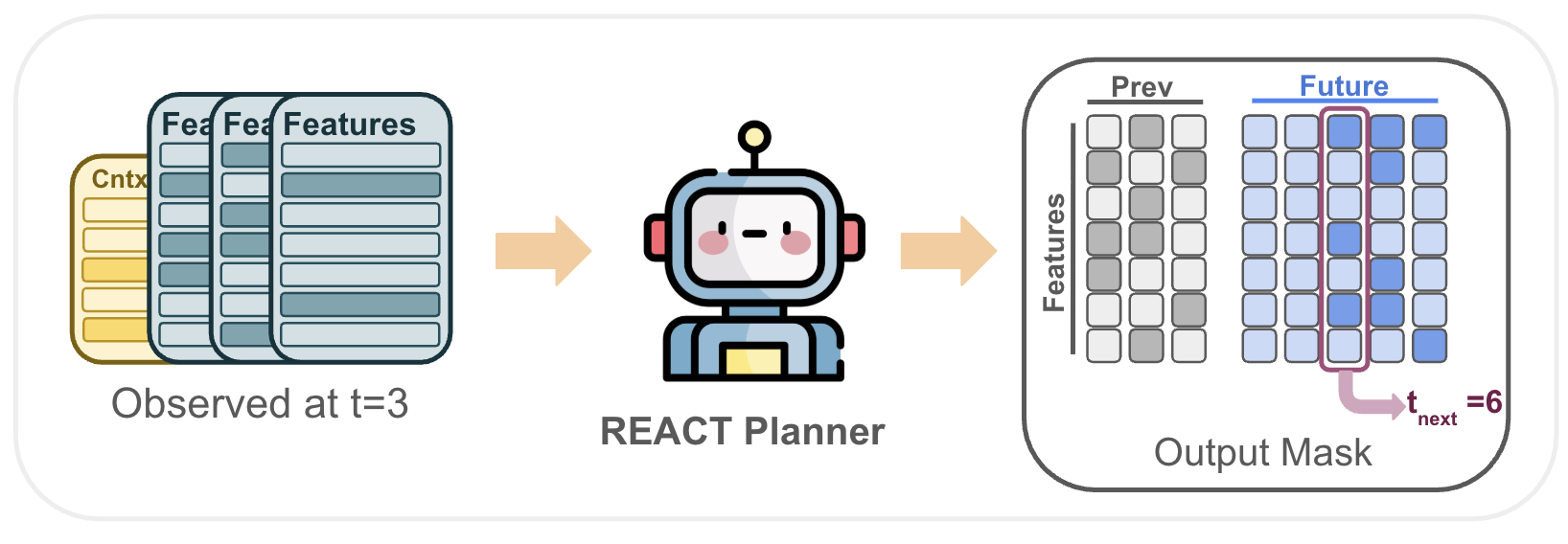}
    \caption{{REACT Longitudinal planner.} Given onboarding context and longitudinal measurements up to t, the planner outputs a binary acquisition mask over future feature--time pairs. The left portion corresponds to previously acquired measurements (unused); the right specifies the future acquisition plan under the cost-benefit objective. The earliest selected future time defines $t_{\mathrm{next}}$.}
    \label{fig:React_planner}
\end{figure}
In the following, we will describe how the variable acquisition process in EMA studies can be cast in the LAFA framework. A glossary of terms is provided in Table \ref{tab:glossary}. The goal of LAFA is to obtain a time-varying policy that takes as input the history up till time $t$ and produces an acquisition plan for the future (see Figure  \ref{fig:React_planner}). To be precise, let us define an acquisition mask $\bm{m}_s \in \{ 0, 1\}^{d_s}$ that determines which of the demographic variables are to be selected. Note that this mask does not depend on the individual $i$. We can also define an acquisition mask $\bm{m}_{x_t}^{(i)} \in \{ 0, 1\}^{d_x}$ that determines which temporal variables are to be selected at time $t$ for person $i$. Unlike $\bm{m}_s$, these masks are personalized because they depend on the actual values of the selected demographic variables and the history of previously observed values (see below). The entire acquisition plan up till time horizon $T$ is then the collection of masks $\{\bm{m}_s, \{ \bm{m}_{x_{1:T}}^{(i)} \}_{i=1}^N \}$, where $\bm{m}_{x_{1:T}}^{(i)}$ is the set of masks of a given window length for all $t$ for person $i$. Our goal is to develop a method that will sequentially produce each of these binary mask vectors based on the dataset $\mathcal{D}$. 

To do so, let us first note that the intermediate masks $\bm{m}^{(i)}_{x_t}$ are related to the initial mask $\bm{m}_s$ and the preceding masks $\{ \bm{m}^{(i)}_{x_{t'}} \}_{t' < t}$ in a particular way. Let $\bm{x}^{(i)}_t$ be the $t^{th}$ column of $\bm{X}^{(i)}$. Consider the masked temporal history 
\begin{align}
    \mathcal{H}_t^{(i)} = (\tilde{\bm{x}}^{(i)}_1, ..., \tilde{\bm{x}}^{(i)}_t, \bm{0}, ..., \bm{0}) \in \mathbb{R}^{d_x \times T}
\end{align}
where $\tilde{\bm{x}}^{(i)}_t := \bm{x}^{(i)}_t \odot \bm{m}^{(i)}_{x_t}$ is the masked vector of temporal variables for person $i$ at time $t$ and $\odot$ represents element-wise multiplication. Similarly, define $\tilde{\bm{s}}^{(i)}_t := \bm{s}^{(i)}_t \odot \bm{m}_{s}$ as the masked vector of demographic variables for person $i$. The next acquisition time and mask is given by a {\it planner} $\pi$ that maps the masked history and time index to the next acquisition time and mask. Specifically, 
\begin{align}
    \pi_{\theta}(\mathcal{H}_t^{(i)}, \tilde{\bm{s}}^{(i)}_t, t) \rightarrow (t_{\text{next}}^{(i)}, \bm{m}^{(i)}_{x_{t_{\text{next}}}}) \quad \text{or} \quad \pi_{\theta}(\mathcal{H}_t^{(i)}, \tilde{\bm{s}}^{(i)}_t, t) \rightarrow \varnothing
\end{align}
where $\bm{\theta}$ is the set of parameters used to parameterize the planner. $\varnothing$ here indicates the plan output form the planner $\pi$ is not acquiring any features in any of the future timesteps (i.e. stop collecting data). This planner is then iterated sequentially through time to yield each of the masks $\bm{m}^{(i)}_{x_t}$. For timesteps that are skipped over or missing, the acquisition mask is simply a vector of zeros. Together with the initial mask, this iterative process produces the collection of masks that is desired across all timesteps. 

Obviously, an arbitrary sequence of acquisition masks might not be the most useful in practice (e.g. the trivial acquisition plan where all masks are set to zero). Therefore, we need a way to decide on the optimality of a particular set of acquisition masks. We do this by introducing a prediction model parameterized by $\phi$:
\begin{align}
    \widehat{y}^{(i)}_t = f_{\phi}(\mathcal{H}_t^{(i)}, \tilde{\bm{s}}^{(i)}_t, t) \in \Delta_{C-1}
\end{align}
which takes the history and timestamp as input to produce a forecast of the outcome variable at time $t$. In the above, $\Delta_{C-1}$ is the $(C-1)$ probability simplex. These predicted class probabilities can then be compared against the actual outcome variable and their difference provides a signal that we can use to score the utility of a given acquisition plan. 

In summary, there are three interacting modules that are combined to produce the desired acquisition plan:
\begin{enumerate}
    \item {\bf Context Selector} $\bm{\alpha} \in \mathbb{R}^d_s$: a learned vector that parameterizes the binary context mask $\bm{m}_s$ at $t=0$. 
    \item {\bf Longitudinal Planner} $\pi_{\theta}$: a neural network that maps the history up till time $t$ and produces the subsequent masks for $t' > t$. 
    \item {\bf Prediction Model} $f_{\phi}$: a neural network that maps the history up till time $t$ and produces predicted class probabilities $\widehat{y}^{(i)}_t$. 
\end{enumerate}
A fixed set of parameters correspond to a specific set of personalized, dynamic acquisition plans encoded by the acquisition masks. In the following, we will describe how the parameters of these three components are optimized by minimizing a custom objective function tailored for the EMA setup. 

\subsection{REACT Objective Function}

There are two competing desires in our choice of the acquisition masks. On the one hand, we want to obtain masks that maximise our ability to predict the categorical criterion of interest. At the same time, we also want our acquisition scheme to minimize participants' burden. To this end, we train our model to directly optimize this accuracy-cost tradeoff. Let $\bm{c}_s \in \mathbb{R}_{+}^{d_s}$ be a vector of costs denoting the cost of acquiring each of the demographic variables, and let $\bm{c}_x \in \mathbb{R}_{+}^{d_x}$ denote the vector of costs for acquiring each of the temporal features. These costs are defined in an abstract manner and can be used to reflect the amount of effort, time or risk for a participant to provide a response to the variable in question. Using this definition, the cost of a given acquisition scheme for person $i$ is given by:
\begin{align}
    \text{Cost}(\{\bm{m}_s, \{ \bm{m}_{x_{1:T}}^{(i)} \} \}) = \bm{c}_s^\top \bm{m}_s + \sum_{t=1}^T \bm{c}_x^\top \bm{m}_{x_t}
\end{align}
The overall loss function is then a sum of a typical cross-entropy classification loss with the cost of the acquisition scheme:
\begin{align}
    \sum_{i=1}^N \bigg( \sum_{t=1}^T \mathcal{L}_{\text{cross-entropy}} (\widehat{y}^{(i)}_t, y^{(i)}_t) + \lambda \text{Cost}(\{\bm{m}_s, \{ \bm{m}_{x_{1:T}}^{(i)} \} \}) \bigg)
\end{align}
where the quantities $\widehat{y}^{(i)}_t$ and $\{\bm{m}_s, \{ \bm{m}_{x_{1:T}}^{(i)} \} \}$ are implicitly functions of the model parameters $(\bm{\alpha}, \bm{\theta}, \bm{\phi})$. In the above, $\lambda$ is a prediction-cost tradeoff hyperparameter. In practice, we utilize a self-iterative training process rolling out the policies own states for training, similar to DAgger-style iterative imitation leaning \parencite{Ross2010ARO}, but driven by direct optimization with ground truth labels instead of imitation targets (see details in the Supplementary Materials). Moreover, the discrete masks in the loss are replaced by a smooth approximation using Gumbel-Sigmoid relaxation \parencite{Jang2016CategoricalRW,maddison2016concrete}. A self-contained description of this approximation can be found in the Supplementary Materials.

\section{TREACT: Time-indexed Tree Distillation from Black-Box Neural Network Policy}
% \textcolor{red}{@Bing; I tried to motivated the Tree policy in this following paragraph, could you make adjustments to the wording for a psych audience? }

% \todo{Add sentence to pitch what TREACT is doing}
% done --- this is what katie wanted
In this section, we will develop TREACT, a method that produces interpretable  acquisition policies using tree-based methods. The key motivating problem is that REACT yields an accurate but opaque policy: the acquisition decision is a forward pass through a neural longitudinal planner $\pi_\theta$. This is a black-box process. The mapping between history to the next acquisition mask is a function parameterized by an opaque neural network, which can be hard to interpret. More importantly, a black-box acquisition policy can be difficult to defend, justify or amend when there are legal or ethical considerations at stake. In this sense, a black-box neural policy can be fairly unattractive. 

% This can be clunky to work with in practice. If a researcher wishes to utilize the policy produced by REACT, they will need to load the entire trained neural network into a computer and feed in the entire acquisition history before they can extract a mask vector to know which variables to acquire next. Such a black-box process can be challenging to work with, especially if the number of features available is large. 

\begin{figure}
    \centering
    \includegraphics[trim={0.5cm 0.5cm 0.5cm 0.5cm}, width=0.9\linewidth]
    {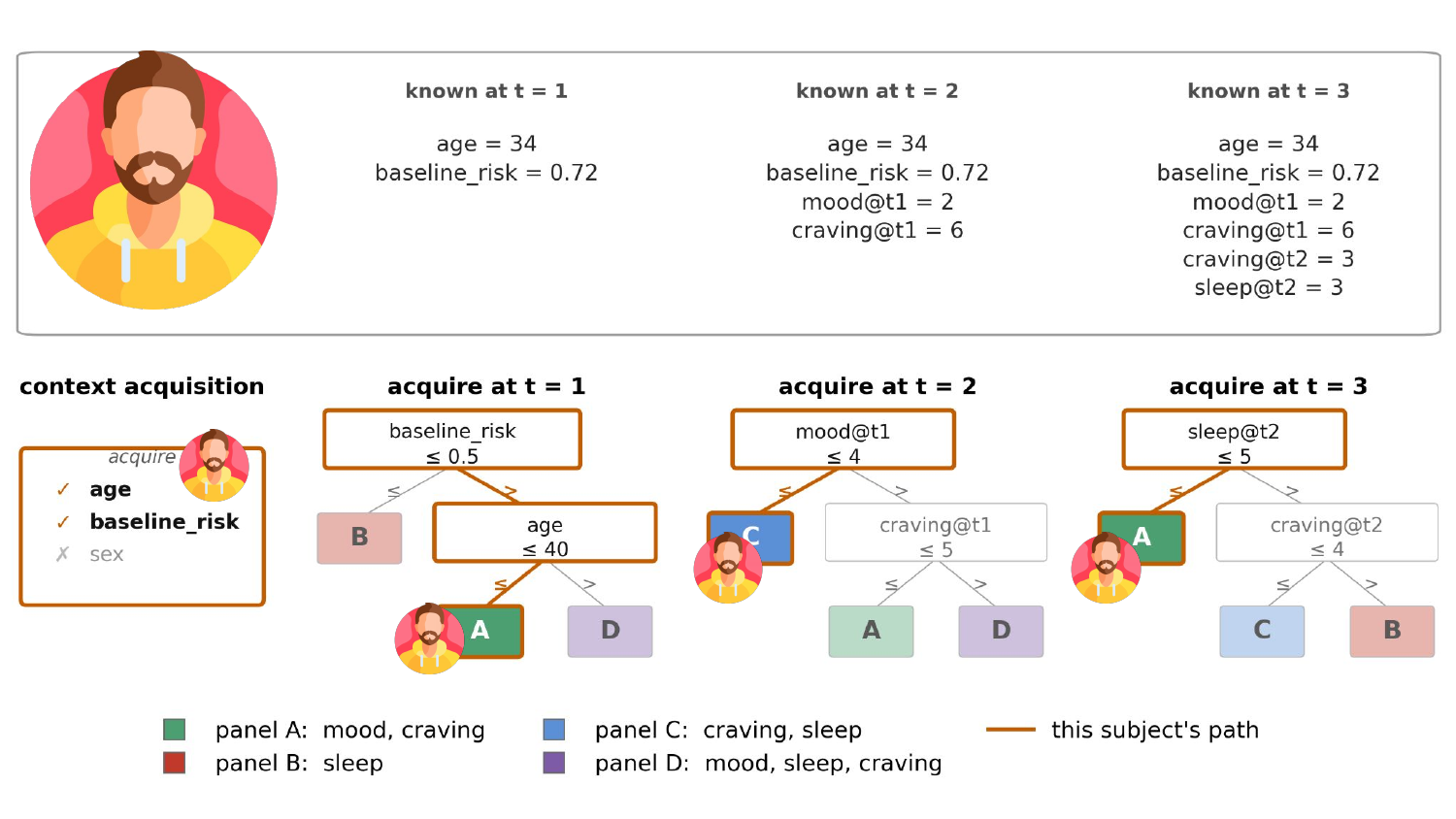}
    
    \caption{The tree as a protocal with a toy example. Here we have a subject with feature values and we trace through the subject's path on the tree. The orange borders and branches are the subject's path. Note that `sex' is never acquired in this tree policy, but is included in the diagram because the data that TREACT is trained on contains the `sex' variable.}
    \label{fig:subject_tarce_through}
\end{figure}

TREACT gets around this opacity by distilling the neural policy into a tree policy that domain experts can execute by hand. Consider the toy example represented in Figure \ref{fig:subject_tarce_through}. At onboarding, we only need to collect age and baseline risk; the sex variable does not need to be obtained. Given this information, we can then make decisions from the tree at the first measurement period. Specifically, because their baseline risk is greater than 0.5 and they are less than 40 years old, we decide to acquire a panel of variables named `mood' and `craving'. At the next measurement period, we can then move on to the second tree and check the value of their mood acquired at the first measurement occasion. Having noted that they scored less than 4 on the mood variable, we then proceed to ask them about their cravings and sleep behavior. We can similarly proceed to read off the third tree at the third measurement occasion. Instead of a black-box function, the distillation process gives us one shallow decision tree per measurement period. At a given acquisition time, the leaves of the corresponding tree lead us to a particular acquisition pattern produced by REACT (i.e. REACT produces a finite number of acquisition patterns). Such a transparent acquisition process represented by decision trees allow researchers to clearly articulate the acquisition process. There are also several practical ancillary benefits of a tree-based policy: a tree policy can be easily audited, put in a protocol document, or be amended by domain experts. This motivates the need for a tree-based approximation of a neural policy in the context of EMA research.

% These patterns can be represented as a small codebook that a researcher can execute directly by following the tree. Consider the toy example represented in Figure \ref{fig:subject_tarce_through}.

% At onboarding, we only need to collect age and baseline risk; the sex variable does not need to be obtained. Given this information, we can then make decisions from the tree at the first measurement period. Specifically, because their baseline risk is greater than 0.5 and they are less than 40 years old, we decide to acquire a panel of variables related to mood and craving. At the next measurement period, we can then move on to the second tree and check the value of their mood acquired at the first measurement occasion. Having noted that they scored less than 4 on the mood variable, we then proceed to ask them about their cravings and sleep behavior. We can similarly proceed to read off the third tree at the third measurement occasion. Such a transparent acquisition process represented by decision trees allow researchers to clearly articulate the acquisition process. There are also several practical ancillary benefits of a tree-based policy: a tree policy can be easily audited, put in a protocol document, or be amended by domain experts. This motivates the need for a tree-based approximation of a neural policy in the context of EMA research. 

% We can use the tree policy to acquire useful features for a subject with certain features values with out running any model. INSERT EXAMPLE WALKTHROUGH. 

\subsection{Executing the Distilled Protocol}
% NOTE TO YUNNI: I actually think Katie just wanted the one sentence that I added at the start of this section (see above).
\begin{figure}[h!]
    \centering
    \includegraphics[width=0.9\linewidth]{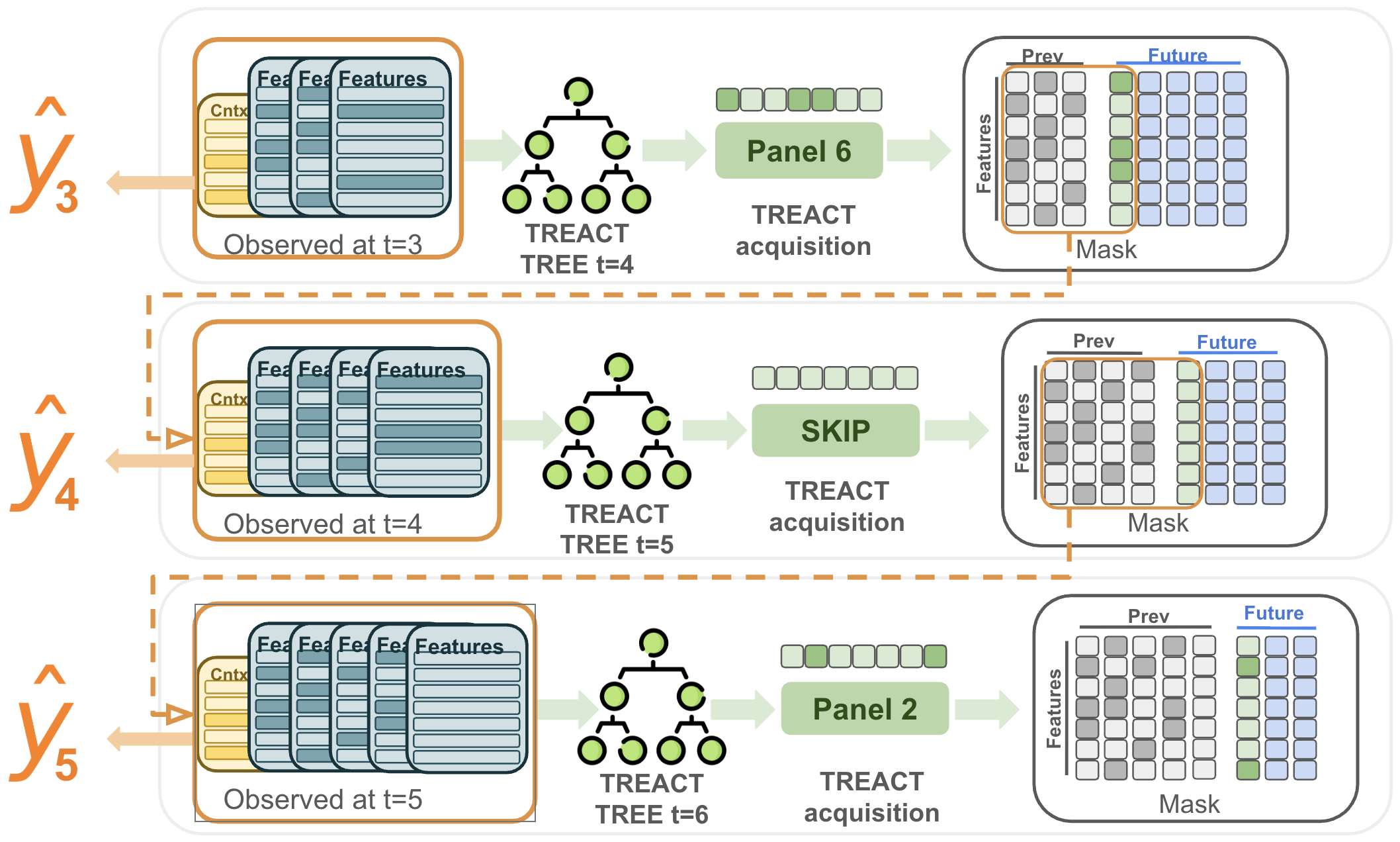}
    \caption{Execution of a distilled TREACT protocol across three consecutive decision occasions for one participant. At each occasion, the accumulated context and temporal history are passed to the corresponding tree, whose leaf returns an element of the codebook panel $\mathcal{P}$ to acquire for t. Panels 6 and 2 are acquisition patterns, and the mask panel on the right shows the resulting acquisition applied to the current occasion (previously acquired items in grey, the current acquisition in green, occasions not yet reached in blue). The middle occasion returns the "SKIP" action, so no items are administered, no cost is incurred, and the history passed forward is unchanged. A forecast $\widehat{y}_t$ is produced at every time step, including the t=4 on which the protocol waits, and is generated by the TREACT prediction model $f_{\tilde{\phi}}$, which is finetuned classifer $f_\phi$ from REACT.}
    \label{fig:treact_execut}
\end{figure}
Before describing how the trees are fit, it is useful to see how the resulting protocol is executed in the field, since this is what a researcher administering the study would actually do. Figure \ref{fig:treact_execut} traces three consecutive decision occasions for a single participant. At each occasion $t$, the researcher holds the participant's masked history $\mathcal{H}_{t-1}^{(i)}$ together with the masked onboarding vector $\tilde{\bm{s}}^{(i)}$, consults $\mathrm{Tree}_t$, and follows the branches down to a leaf. Each leaf carries one element of the codebook $\mathcal{P}$, which is either an acquisition panel (i.e. a mask mapping to a subset of the features being acquired) or the SKIP/STOP action $\varnothing$. No neural forward pass is required at any point in this loop; the entire acquisition decision is a sequence of comparisons against thresholds that are printed in the protocol document.
 
Two important points of the loop are worth stating explicitly. First, the history grows only when an acquisition actually occurs. In the middle row of Figure \ref{fig:treact_execut}, $\mathrm{Tree}_5$ returns the "SKIP" action, so no items are presented to the participant at that occasion and the history carried into the next occasion is unchanged from the previous one. The acquisition cost incurred at that occasion is therefore zero, which is precisely the mechanism through which the policy purchases reductions in participant burden. Second, a forecast $\widehat{y}_t^{(i)}$ is produced at every occasion regardless of whether anything was acquired, predicting on the previously acquired history only. 

% On occasions where the tree waits, the prediction is formed from the existing history alone, so the forecast is available but is based on existing information.
A final point concerns which components of the trained model are needed at deployment. The distillation replaces the longitudinal planner $\pi_{\theta}$ with the sequence of trees, and it is this replacement that makes the acquisition protocol auditable and executable by hand. The classification model $f_{\phi}$ from REACT is finetuned with training sample rollout on the trees to get a predictor $f_{\tilde{\phi}}$ that saw TREACT's acuiqistions patterns at training time. Generating the forecast $\widehat{y}_t^{(i)}$ therefore still requires evaluating the neural predictor. It is worth distinguishing this goal from interpretable classification.
While interpretability has been well studied in the Machine Learning community \parencite{Guidotti2018ASO, Rudin2021InterpretableML}, interpretability has not been explored much in the LAFA setting. Therefore, the goal of our method is to distill a tree-based LAFA policy for acquisition, and any of the standard approaches to interpretable classification could be substituted for $f_{\tilde{\phi}}$ without altering the distillation procedure described below.
% Our scope here is the acquisition policy: interpretable classification is a well-developed problem with a large existing literature \parencite{Bertsimas2017OptimalCT, Letham2015InterpretableCU, Rudin2018StopEB}, whereas interpretable \emph{acquisition} policy is not, and the acquisition policy is the component a researcher must write into a protocol document and administer. 
% whose leaves name panels drawn from a small codebook. In a clinical setting the acquisition policy isn't deployed as a model, it's deployed as a protocol someone follows. A neural planner on the other hand requires the trained network, its predictor, and the full acquisition history at inference. A visit-indexed tree is a page: at baseline read the on boarding context; at visit 3, if item\_2 $\leq$ 0.54, order panel P0. That's a form a domain expert can easily execute, audit, and put in a protocol document. In static feature selection, a decision decision can be explained via feature importance. Here in the LAFA setting are T sequential decisions, each conditioned on what earlier ones revealed, so a saliency map or feature-importance ranking doesn't capture the object, the policy is a branching plan over time, and a per-visit tree with routing between visits is a natural representation of exactly that. 

\subsection{Time-Aware Recursive Partitioning for TREACT Tree Fitting}

% {\color{red} Question for Yunni: how are we creating the matrix if we are running REACT where the number of timepoints exceed the window size? e.g. if we have 13 timepoints but a window of 10? Do we just use the first 10, or concatenate all the first measurement of the rolling windows to get 3N rows?}
% {
% \color{blue} if you are talking about test  time, we can just split into 2 sliding windows for 13. I think what matters the most is at a deployment time we need to state the limitation that we can only take up to a limited number days of history with this architecture. In our case, we trained with 10 days so we can take 9 days of history and after that we start to loose the begining of the sequence.
% }

We will rely on recursive partitioning \parencite{strobl2009introduction, breiman1984classification} to develop such a tree policy. Recursive partitioning is an algorithm that allows us to construct decision trees. The input to this algorithm is a feature matrix $\bm{F}$ where the rows are participants and the columns are features. Each row is paired with a class label, which can be collected in an outcome vector $\bm{O}$. The algorithm then (recursively) partitions the rows into different subsets based on the values of particular columns of $\bm{F}$. The choice of columns to partition on correspond to the decision nodes in a tree, and each leaf in the tree correspond to a subset of rows. Predictions in each leaf partition can then be made via a majority vote of the labels.

In our context, recall that the REACT policy is our `teacher' and the TREACT policy is the `student'. The goal here is to distill (from teacher to student) the black-box neural policy from REACT into a transparent tree-based policy. We do this by (a) creating the labels $\bm{O}$ from the REACT model and (b) training the tree to best predict these labels given the features available. However, there is some subtlety in how recursive partitioning has to be carried out. Each tree at time $t$ (i.e. $\text{Tree}_t$) should only take as input features variables that have been acquired at previous timesteps $t' < t$ based on the TREACT policy, and not the teacher policy. These constraints can be met by growing these trees in a time-aware manner, the details of which we will describe below.

We will begin by fixing some notation. Let $\bm{X}_t \in \mathbb{R}^{N \times d_x}$ be the matrix of temporal features obtained at time $t$ and let $\bm{s} \in \mathbb{R}^{N\times d_s}$ be the matrix of onboarding features. By running the REACT policy on the training data, we obtain a mask matrix $\bm{M}_t^{\text{REACT}} \in \{0,1\}^{N \times d_x}$ at each time step $t$. Note that we set the onboarding mask for our tree policy to be the same as that in the REACT policy. Since there are only a finite number of possible distinct mask vectors, we can reduce this into a codebook of patterns $\mathcal{P} = \{P_1,\dots,P_K\}$ where each $P_i$ is one particular mask pattern. Although the total number of possible patterns is massive, the number of patterns produced by REACT in practice is a small subset. Accordingly, we can collapse $\bm{M}_t^{\text{REACT}}$ into a $N \times 1$ categorical vector $\bm{O}^{\text{REACT}}_t$. Each element of this vector belongs to $\mathcal{P} \cup \{\varnothing\}$, where $\varnothing$ denotes that no acquisitions will occur at this time step, which could be a "SKIP" or "STOP" action. Therefore, the total number of possible labels is $|\mathcal{P}|$ + 2. These pattern labels will be our outcome of interest when running recursive partitioning. Quantities bearing the TREACT in the superscript are similarly defined. Moreover, let $\bm{E}^{\text{TREACT}}_t \in \{ 0, 1\}^{N \times 1}$ be a binary vector that tracks if a person's acquisition process has halted by time $t$. 
\begin{equation*}
\left[\bm{E}_t^{\mathrm{TREACT}}\right]_{i,1}
=
\mathbb{I}
\left\{
\exists\, k\leq t:
\left[\bm{O}_k^{\mathrm{TREACT}}\right]_{i,1}
=
\mathtt{stop}
\right\}.
\end{equation*}
\subsubsection{Learning $\text{Tree}_{1}$}

At $t = 1$, we want to learn a tree using standard recursive partitioning algorithms. To do this, we supply it with a $N \times q_1$ matrix of features $\bm{F}_1$, where $q_1$ is just the number of onboarding features. This is accompanied by the vector of outcomes $\bm{O}^{\text{REACT}}_1$. Since we only have access to the onboarding features in $\bm{s}$ at onboarding, we can simply set $\bm{F}_1 = \bm{s}$. The tree splits are chosen based by minimizing Gini impurity \parencite{nembrini2018revival}. In essence, we are training a tree model that will best mimic the outputs of REACT at $t=1$ using the onboarding information only.

\subsubsection{Learning $\text{Tree}_{t}, t > 1$}

At time $t > 1$, we will have access to the previously learned trees from time $\{t-1, t-2, ..., 1 \}$. We will use these trees to construct an expanded feature space $\bm{F}_t$ to predict the outcomes $\bm{O}^{\text{REACT}}_t$. Note that the feature matrix has to be constructed in such a way that the partitioning uses only variables that were acquired on previous timesteps by the tree policy. If we do not take into account such temporal dependence, we can end up with an incoherent tree policy that makes decisions based on variables that it never acquired. 

We enforce this constraint by sequentially fitting tree models to distill from the REACT neural policy. This is done in a few steps. Firstly, we will use the previous trees $\{\text{Tree}_k \}_{k < t}$ to get predicted pattern labels for each timestep $\{ \bm{O}^{\text{TREACT}}_k \}_{k < t}$, which is also used to update $\bm{E}^{\text{TREACT}}_{t-1}$. These pattern labels can be expanded into the mask matrices $\{ \bm{M}_k^{\text{TREACT}} \}_{k < t}$. These masks are then applied to the data matrices via $\widetilde{\bm{X}}_t = \bm{X}_t  \odot \ \bm{M}_t^{\text{TREACT}}$. We then stack these masked matrices columnwise:
\begin{align}
    \bm{Z}_t := [ \bm{s} \ \widetilde{\bm{X}}_1 \  \widetilde{\bm{X}}_2 \ \cdots \ \widetilde{\bm{X}}_{t-1} ]
\end{align}
This corresponds to an expansion of the feature space to utilize features acquired from all previous timesteps. A few processing steps are then applied to $\bm{Z}_t$: we replace all elements where the mask is zero (i.e. non-acquired variables) with `NA's, drop columns that are entirely `NA's and remove rows with value equal to 1 in $\bm{E}^{\text{TREACT}}_{t-1}$.  We then let $\bm{F}_t$ be the resulting feature matrix. These steps ensure that the partitioning algorithm to grow $\text{Tree}_t$ uses only variables that it has acquired for some individuals at some point in its history. The `NA's are handled by a trinary split \parencite{Zakrisson2023TrinaryDT}. Together with the $\bm{O}^{\text{REACT}}_t$ as the outcome, we can now run a standard recursive partitioning algorithm to produce a valid tree policy. Note that the labels come from REACT, but the masking of variables in the feature matrix used for partitioning come from TREACT. This procedure is similar to behavioral cloning techniques used in the machine learning literature \parencite{galashov2022data}, but is tailored to the active acquisition setup that we have here.  The code used to implement the method is available via the OSF link \href{https://osf.io/uh7b6/overview?view_only=f16e9bd5d4d1494ca349352bdac5cc17}{here}.

\section{Simulation Study}

% Emphasize that this is not meant to capture all the intricacies of EMA 

As proof of concept, we conducted a simulation study to assess how well our LAFA method can recover important variables in a dynamic setting. Since EMA studies can vary widely in their complexity and design, we opted to focus on a relatively simple example to demonstrate how our LAFA method compares with typical static variable selection strategies. 

\subsection{Simulation Setup}

% We evaluated TREACT, a controlled longitudinal simulation designed to test whether an acquisition policy can recover a conditional measurement rule rather than merely identify globally predictive variables. Each participant has four baseline variables: one \emph{pointer} variable and three independent noise variables. 

To evaluate TREACT, we designed a toy simulation that is meant to test if we can recover a conditional acquisition rule rather than merely identify globally predictive variables. This simulation is meant to be a proof-of-concept rather an exhaustive evaluation of the algorithm under reaslitic EMA data conditions. In practice, for lengths of observations $T$ (e.g., $T>60$) one would window or segment each series so that (1) the results remain interpretable rather than have a tree for each timepoint and (2) the policy can scale to time series of varied lengths, which is often seen in empirical data across individuals. For this setup, we consider $T = 8$ decision times, which represent reasonable window sizes that may be used for applied data.  For each person, we generated four onboarding variables: three of which are noise variables and one being a {\it pointer} variable. The pointer assigns the participant to one of three equally likely groups. At each decision time, the participant has six candidate temporal items, but only two are informative: participants in groups $1$, $2$, and $3$ require items $\{1,2\}$, $\{3,4\}$, and $\{5,6\}$ respectively. The binary outcome is then generated from a logistic model whose linear predictor is the sum of the two group-specific items. Therefore, the minimum optimal cost to acquire all the informative features is 2 at each of the 8 time step plus the pointer in the onboarding phase, adding up to 17 (assuming that acquisition of each item has unit cost). The simulation is constructed so that the relevant variables cannot be identified by a global static policy. Instead, the optimal policy first acquires the pointer and then, conditional on its value, acquires only the corresponding pair of longitudinal items at each subsequent decision time.

All onboarding and temporal variables are drawn from the same centered uniform scale, and the pointer differs from the three onboarding noise variables only because its value determines which pair of temporal variables is useful. We generated data for 4,000 windows used for training, and another 1,000 windows for testing. Performance is evaluated over the resulting 1000 held-out windows over each of the 8 time steps. The simulation was repeated for a total of 5 Monte Carlo runs. 

\subsection{Model Setup}
\subsubsection{REACT}
We trained REACT at five cost-penalty values, \[\lambda \in \{0.0005,\;0.001,\;0.00195,\;0.0025,\;0.006\},\]
to trace the cost--performance frontier. Each setting was trained with five random seeds, and the reported performance is the mean across seeds with seed-level uncertainty shown in the frontier plots. The prediction network was a multilayer perceptron with a 64-unit hidden layer and dropout rate $0.1$. The acquisition planner used a three-layer MLP with hidden widths $(256,128,64)$ and a 64-dimensional time embedding.

\subsubsection{TREACT}
For every REACT checkpoint at each $\lambda$ and seed, we distilled a sequential ternary decision-tree policy with maximum depth of 2. At each decision time, the corresponding tree was fit only to information acquired before that time, and its leaves can emit an acquisition panel, \textsc{Wait}, or \textsc{Stop}. TREACT performance was evaluated on the same held-out test participants as its corresponding REACT teacher; results were then averaged over the five matched seed pairs.

\subsubsection{Baselines}
We compared REACT and TREACT with random acquisition and Lasso baselines over a range of acquisition budgets. The random baseline samples available feature groups without using participant information. The Lasso baseline uses an $\ell_1$-penalized logistic model, with its regularization parameter varied to produce a feature-cost--performance curve. Both baselines were evaluated on the same test split and summarized over their 5 independent runs.

\subsection{Metrics of Interest}
% \todo{jaccard formula}
To measure the quality of an acquisition policy, we used AUROC and AUPRC versus cost. A good acquisition policy should be able to acquire the most useful features at a relatively low cost. Therefore, we plot the frontier plots by sweeping over a number of budgets (i.e. $\lambda$, which represents the cost-prediction tradeoff) for each policy to compare their feature utility under different cost constraints. For TREACT, we also rendered the decision trees to visualize if the acquired features are optimal qualitatively.

% {\color{red} @Yunni: Need to insert how we are measuring the success of our method}

\begin{figure}[h!]
    \centering
    \includegraphics[width=0.9\linewidth]{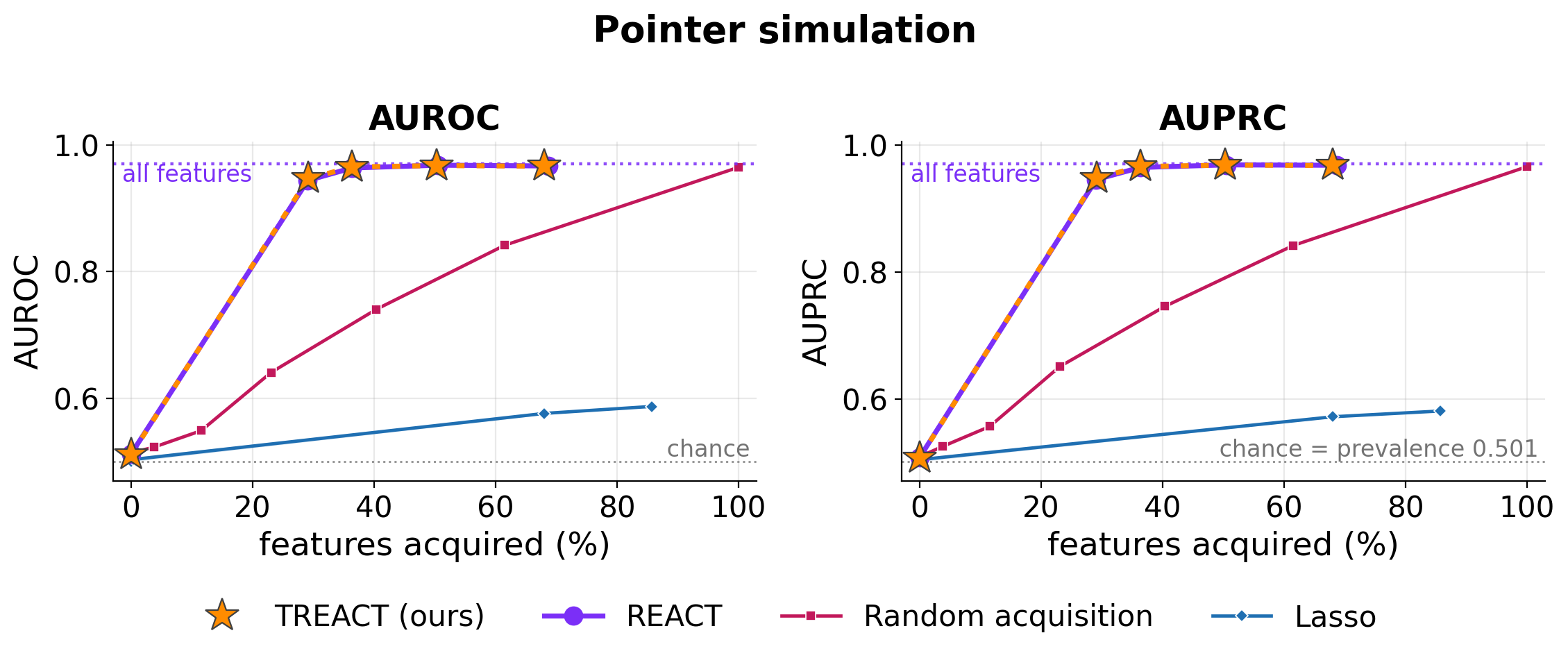}
    \caption{AUROC/AUPRC vs cost for the simulation study. Purple dotted line indicates REACT's classifier performance when given all of the features, acting as a empirical upper bound. The minimum cost for all the features needed to make accurate prediction on all time points is 17 for this simulation }
    \label{fig:sim_frontier}
\end{figure}

% \begin{table}[h!]
% \centering
% \small
% \begin{tabular}{@{}lrrrrrrrr|r@{}}
% \toprule
% Timestep & 1 & 2 & 3 & 4 & 5 & 6 & 7 & 8 & Avg. \\
% \midrule
% % Jaccard & 1.000 & 1.000 & 0.883 & 0.781 & 0.797 & 0.710 & 0.830 & 0.845 & 0.856 \\
% Contains GT & 0.806 & 0.871 & 0.871 & 0.935 & 0.990 & 0.993 & 0.995 & 1.000 & 0.933 \\
% \bottomrule
% \vspace{0.5cm}
% \end{tabular}
% \caption{Acquisition predicted from TREACT protocol against the optimal protocol on the toy simulation, by timestep and averaged over participants. \textbf{Contains GT} is the fraction of participant/timesteps at which the optimal set is a subset of the distilled one. Each value is averaged over the 5 Monte Carlo runs.}
% \label{tab:gt-agreement}
% \end{table}

\begin{figure}[h!]
    \centering
    \includegraphics[width=0.9\linewidth]{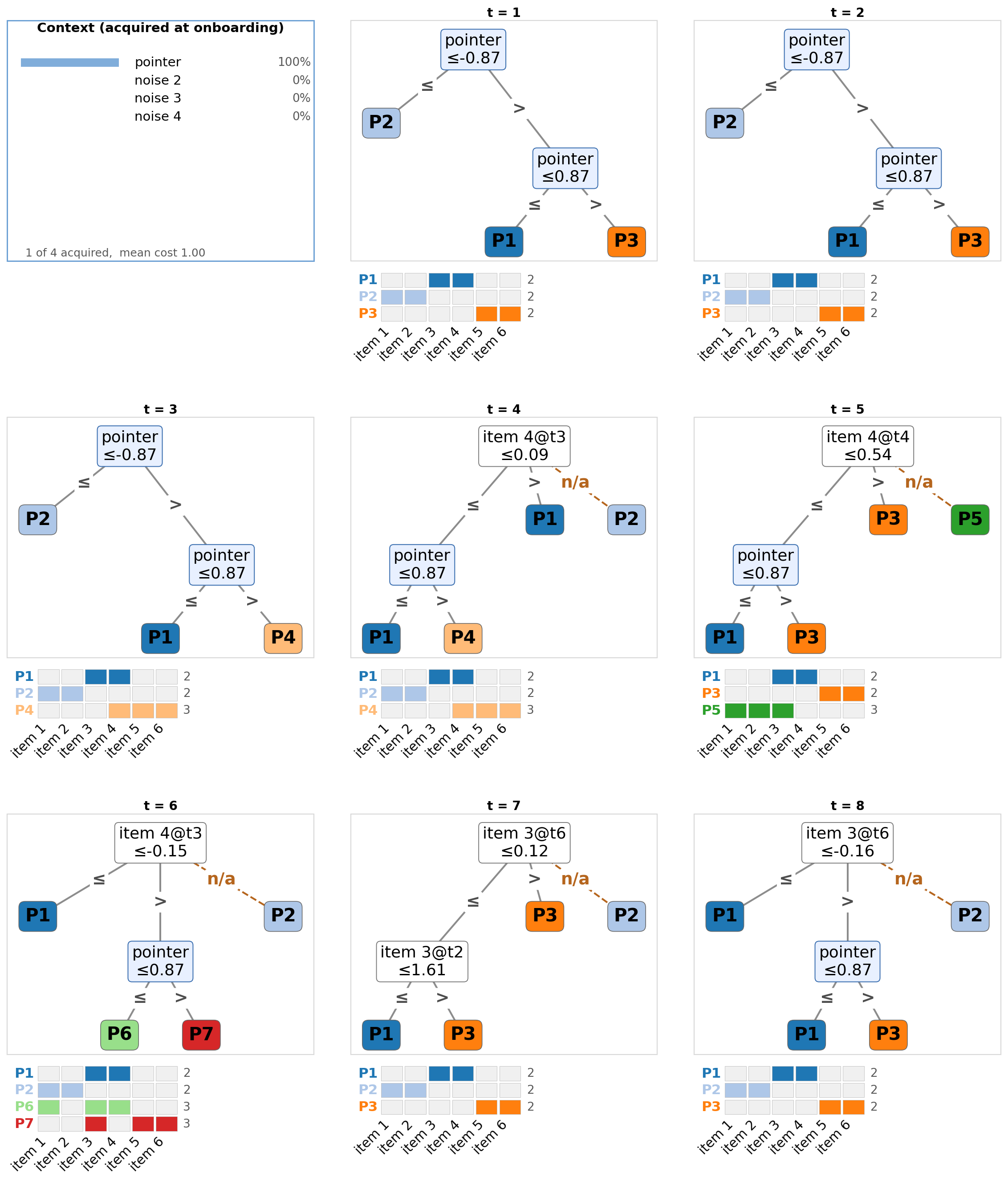}
    \caption{Trees of depth of 2 distilled from the REACT policy closest to the optimal minimum cost in this simulartion. Each panel P is a unique set of features. Each leaf specifies the set of features to acquire at time \(t\), conditional on the decision rules along the path leading to that leaf.}
    \label{fig:sim_tree}
\end{figure}
\subsection{Results}
We compared TREACT with its neural teacher, REACT, on the toy simulation cost/performance frontier (Figure \ref{fig:sim_frontier}). Near the optimal operating point, REACT achieved an AUROC of 0.969, an AUPRC of 0.969, and a mean acquisition cost of 18.42. TREACT distilled this policy into a depth-two, eight-visit protocol with 22 internal split nodes and 11 acquisition panels (i.e., sets of features). The executed tree incurred virtually the same cost (18.47) and, when evaluated with the frozen REACT predictor, achieved an AUROC of 0.968 and an AUPRC of 0.969. Thus, distillation changed AUROC by less than $0.001$ while increasing mean acquisition cost by only $0.04$. TREACT also outperformed random aquisition and LASSO conducted on the concatenated data set with both longitudinal and basleine features across the whole cost range.

\begin{figure}
    \centering
    \includegraphics[width=1\linewidth]{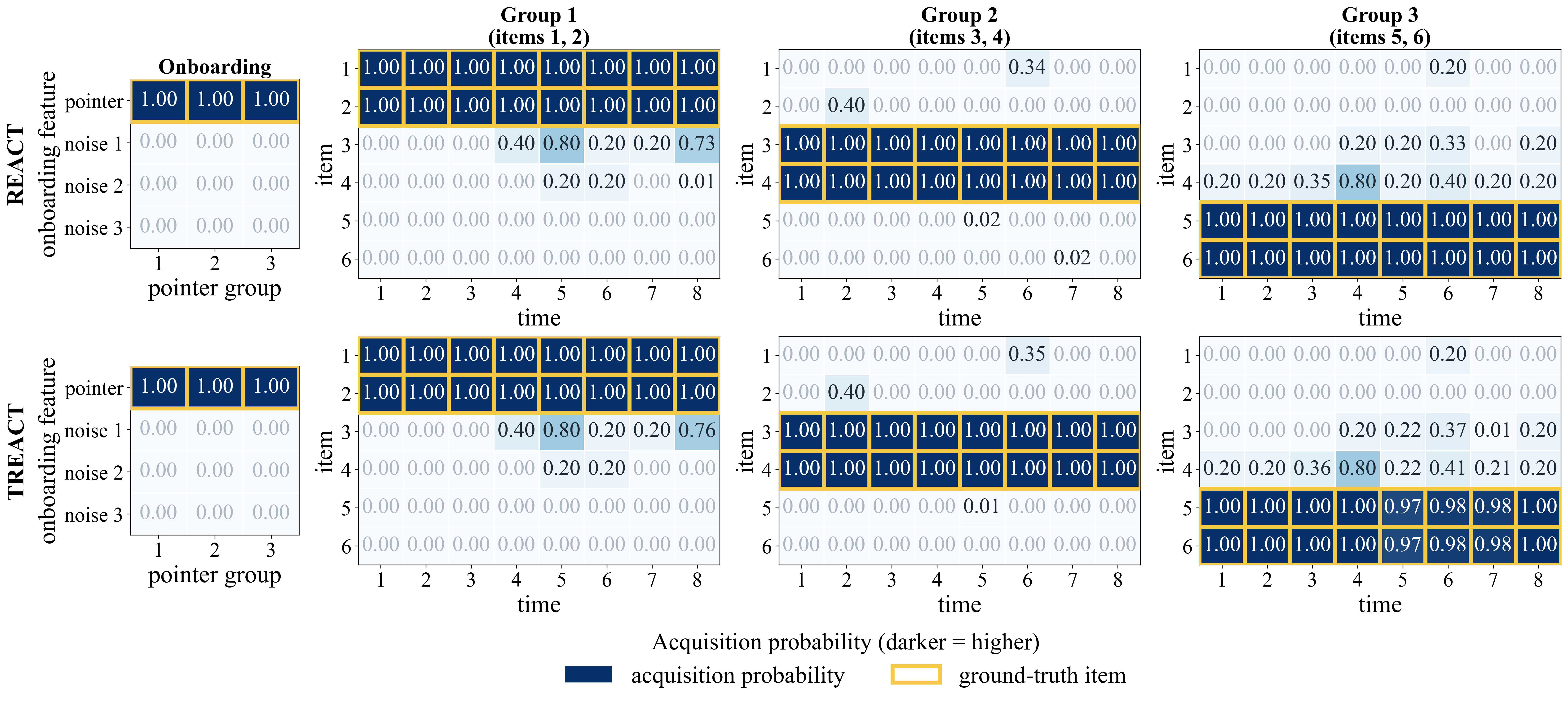}
    \caption{Acquisition heatmap of TREACT abd REACT with the simulation by group averaged over 5 independent runs. Yellow box indicates the ground truth to acquired for each group.}
    \label{fig:sim_heatmap}
\end{figure}

The known data-generating policy provides a direct structural evaluation beyond predictive performance. Fig. \ref{fig:sim_heatmap} shows the acquisition rate of TREACT and REACT along with ground truth annotations on a held-out set. We see here TREACT is able to recover almost all the acquisitions made by REACT. Both TREACT and REACT acquires exactly the pointers and almost all of the ground truth longitudinal features but acquires some longitudinal noise due to the cost slightly bigger than the optimal cost 17. The distilled tree is shown in Figure \ref{fig:sim_tree} which shows a similar pattern. Thus, TREACT converts the published REACT policy into a compact, executable protocol while preserving its cost and predictive performance and recovers the intended pointer-conditional acquisition structure.

\section{Empirical Example}

% {\color{red} @Yunni: I am actually not sure if we want to include ILIAAD. In ILIAAD, we used negative affect as the criterion. But I am guessing that reviewers would instantly question why we did not use any of the other `negative' symptoms as the criterion instead.}

In this section, we demonstrate the use of our method with an empirical EMA example. We will use the "Capturing Habits and Everyday Experiences with Alcohol in Real-time Study" (CHEEARS) dataset, which is an EMA dataset consisting of daily survey responses from 204 young adults \parencite{ringwald2025common}. Each morning, participants are asked to indicate if they drank any alcohol on the previous day. We will use this self-reported drink indicator as our criterion of interest. 

The rest of the data is structured as follows. At baseline ($t = 0$), we collect information on demographics, AUDIT alcohol screening scores, drinking motives (DMQ: social, coping, enhancement, conformity), alcohol consequences (YAACQ: 9 subscales), personality traits (NEO Big Five), and interpersonal problems (IIP: dominance, affiliation, overall problem severity). This provides 35 variables at onboarding. For $t > 0$, participants reported on their affect, drinking urges and expectations, as well as their social experiences each day. This resulted in 149 features measured at each timestep. Because some of the temporal variables belong to the same psychological instrument, we grouped them into 31 groups. Details on the variable groupings can be found in the Supplementary Materials. We assigned a cost of 1 to each group and at acquisition time a group of variables is always acquired together. The dataset is split into training, validation and testing across time with proportions of $(69.5\%, 15.2\%, 15.2\%)$ respectively. The results for the prediction-cost tradeoff are reported on the test partition. 

In this study, each person is followed for up to 120 days. The current model architecture for REACT cannot handle such lengthy sequences. To get around this, we used a rolling window of length 10 and with a stride of 1 to create multiple windows of length 10 from a single person's time series data. REACT is then trained on these windowed versions of the data, similar to the simulated data. We get the cost peformance-trade off curve by sweeping over a range of cost--accuracy tradeoff parameter $\lambda$. The $\lambda$ we used are \{0.0001, 0.0002, 0.0004, 0.0015, 0.002, 0.003\}.

We ran TREACT on the same windowed data used to train REACT. For each setting of the cost--accuracy tradeoff parameter $\lambda$, we took the corresponding trained REACT policy as the teacher and fit a sequence of ten trees, one per within-window decision occasion. The onboarding mask $\bm{m}_s$ was carried over directly from REACT rather than distilled, so the tree sequence governs only the temporal acquisitions. Trees were grown by recursive partitioning with Gini impurity and a maximum depth of 3. Depth was fixed a priori rather than tuned: a depth-three tree fits legibly on a printed page and can be followed by hand during administration, which is the property the distillation is meant to deliver. Following the sequential learning of trees detailed in a previous section: at each occasion $t$, the feature matrix $\bm{F}_t$ was assembled from the onboarding variables and the temporal variables that the \emph{tree} policy had acquired at occasions $t' < t$, with non-acquired entries encoded as missing and handled through trinary splits, while the outcome labels $\bm{O}_t^{\mathrm{REACT}}$ were taken from the teacher. Windows whose acquisition process had halted by occasion $t$ were dropped from the fitting sample for all subsequent trees.

Because acquisition costs are assessed at the group level, the codebook $\mathcal{P}$ is defined over patterns of the 31 temporal groups rather than the 149 individual variables. The trees were fit on the training partition, and all reported cost and performance figures are computed on the test partition, matching the protocol used for REACT. The entire procedure was repeated across 5 independent runs, and Figure~\ref{fig:cheears_frontier} reports means with $\pm 1$ standard error. Predictions were generated by the TREACT prediction model $f_{\tilde{\phi}}$, which is the REACT prediction model $f_{\phi}$ fine-tuned on tree-policy trajectories.

% {\color{red} @Yunni: Whats the dimension of the onboarding features? And the dimension of the temporal features? Probably should report the full set of features in a table in the Supplementary or something.} {\color{blue} }

% {\color{red} @Yunni: I am actually confused how the sliding window of 10-days work... do you mean $\pi_\theta$ takes as input only the last ten days? }

% {\color{red} @Yunni: uh how did we deal with missing data in the data?}

\subsection{Model Architecture}

% {\color{red} @Yunni: We also need to talk about hyperparameters somehow.... and how we at least did some sensitivity analyses...}

We used official implementation from the REACT paper for the longitudinal planner and classifier, where $\pi_\theta$ consists of 3 hidden layers with dimensions $(512, 256, 128)$ and ReLU activations. We trained the planner network with a total of 1,000 batches using a self-iterative training procedure as described in \textcite{qu2026relaxed}. The details are reproduced in the Supplementary Materials. The prediction model $f_{\phi}$ consists of a 3-layer multilayer perceptron producing logits for each class. 

\subsection{Baseline Setup}
\paragraph{Random acquisition.}
We swept through total acquisition budgets\[B \in \{12,\;24,\;40,\;75,\;100\}.\] The selected onboarding variables and available longitudinal slots were sampled uniformly at random, without replacement. The random baseline was evaluated over five random replicates; the frozen REACT predictor scored its acquired measurements. This baseline therefore tests the value of learned acquisition choices while holding the prediction architecture is fixed.

\paragraph{Lasso.}
We fit an $\ell_1$-penalized logistic regression model on the training set using regularization strengths \[C \in \{0.0024,\;0.0030,\;0.0044,\;0.0126,\;0.0200\}.\]Nonzero coefficients defined one global, non-adaptive acquisition protocol: a nonzero coefficient for any feature in a longitudinal group at a particular time caused that group--time slot to be acquired, and nonzero coefficients on onboarding variables caused those variables to be acquired. The Lasso model both selected features and produced predictions.

\subsection{Results}
\subsubsection{Prediction-Cost Tradeoff}
% \todo{REACT, TREACT, Lasso, Random Acq.}
% {\color{red} @Yunni: Insert prediction-cost curves here}
\begin{figure}
    \centering
    \includegraphics[width=0.9\linewidth]{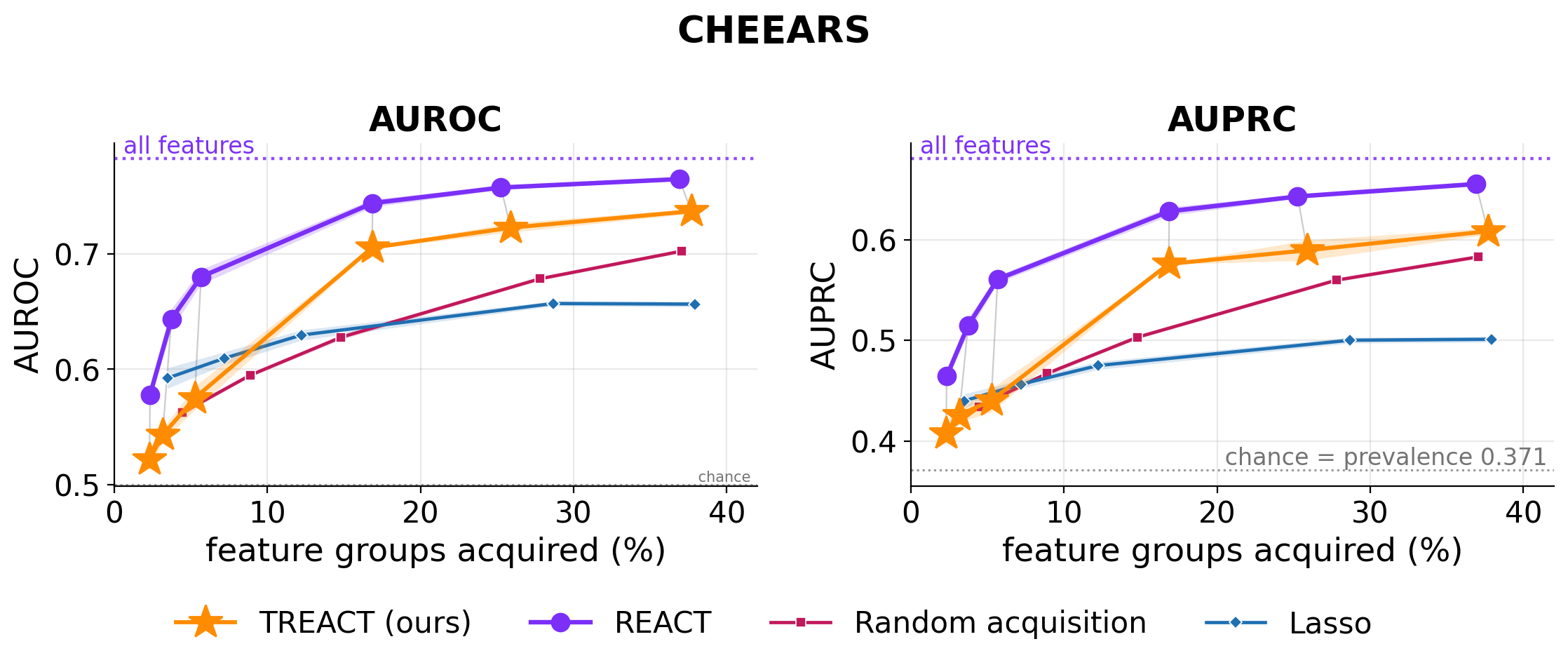}
    \caption{AUROC/AUPRC vs cost of TREACT, REACT and baselines predicting acuiqsitions at unseen times. Purple dotted line is a empirical upper bound from the using neural network classifier of REACT with all features available. Shaded region indicate $\pm 1$ standard error of 5 independent runs.}
    \label{fig:cheears_frontier}
\end{figure}

The cost vs. AUROC/AUPRC frontier (Figure \ref{fig:cheears_frontier}) shows the performance of TREACT using trees with a depth of 3. We obtain this curve by adjusting the cost/performance tradeoff hyperparameter for REACT, and as we decrease this parameter REACT learns a better-performing policy that incurs more cost. The distilled trees from TREACT also follow this general pattern. At a matched cost, TREACT has a lower AUROC/AUPRC than the corresponding REACT policy; this decrease in performance is well known to be trade of for interpretability \parencite{Bertsimas2019ThePO}. A modest reduction in predictive discrimination is incurred in exchange for a fully auditable sequence of feature-acquisition rules. Despite this, TREACT outperformed the 2 baseline methods (random acquisitions and population-level Lasso) in most of the cost range.

Importantly, the role of TREACT is not to improve the cost/performance frontier of REACT. Instead, it provides an interpretable approximation to REACT. At the approximately cost-matched operating point, the tree makes the acquisition process transparent through explicit temporal decision rules while quantifying the associated performance loss. This distinction is important: TREACT is most useful when auditability, protocol review, or restrictions on deploying a neural acquisition policy justify accepting some loss in predictive performance.

% and, in some settings, acquisition efficiency.

\subsubsection{Model Distillation}
\begin{figure}[h!]
    \centering
    \includegraphics[width=0.94\linewidth]{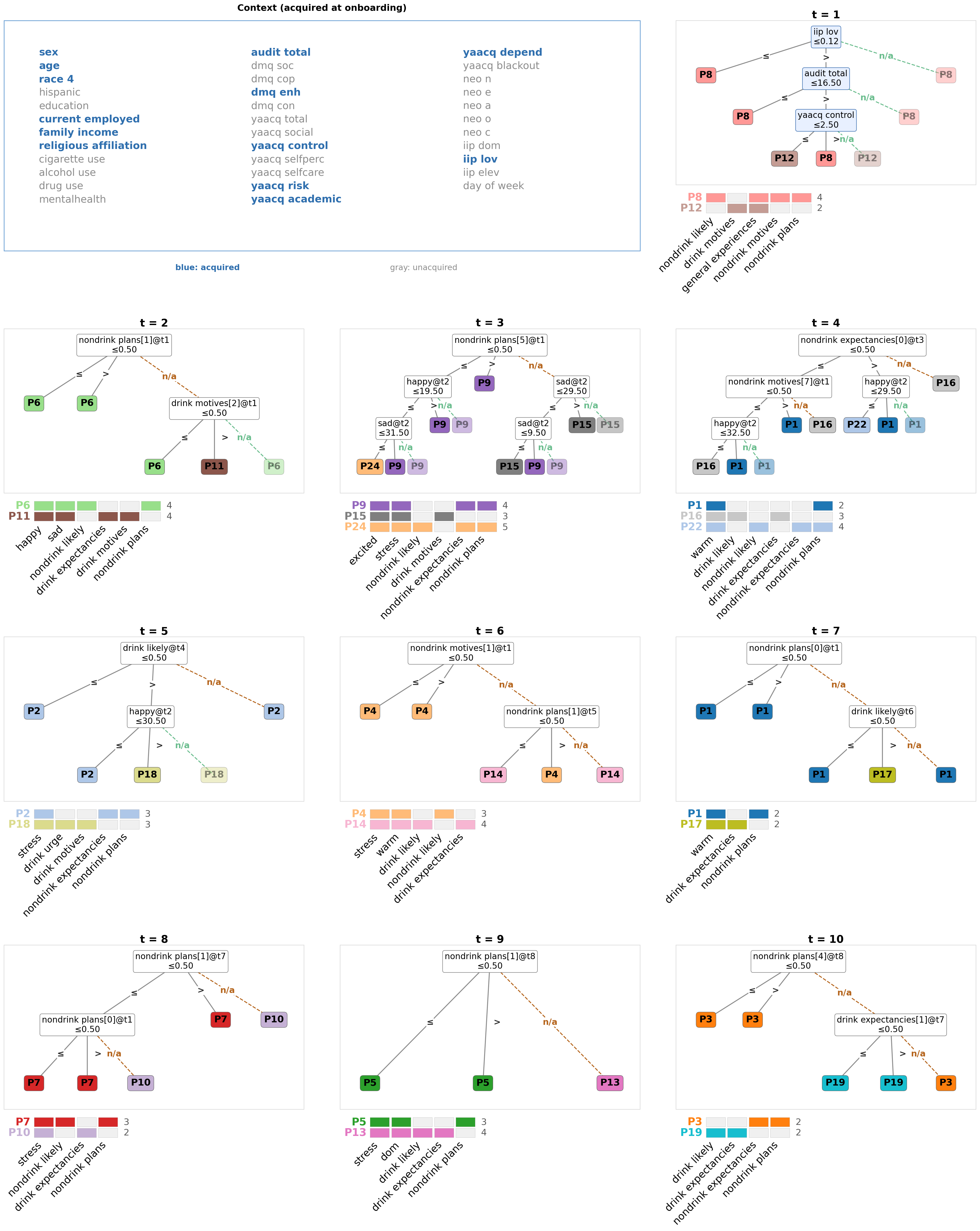}
    \caption{TREACT trees with depth of 3 representing acquisitions for the CHEEARS data. The policy's AUROC is 0.750, AUPRC is 0.643, and average cost is 42.73.}
    \label{fig:cheears_tree}
\end{figure}
In Figure \ref{fig:cheears_tree}, we show a representative tree of depth 3 with AUROC of 0.750 and an AUPRC of 0.643 at a mean acquisition cost of 42.73 from TREACT. Each split in the protocol specifies the currently available measurement, its decision threshold, and the subsequent acquisition panel, making the policy inspectable at every decision time. 

Concretely, at $t = 0$, the TREACT policy identified a subset of 13 demographic variables (e.g. sex, age, family income) that are of relevance in predicting drinking behavior. Several variables were deemed to be relatively uninformative for prediction (e.g. NEO personality scores) after accounting for the acquired variables. At each subsequent timestep, we have a decision tree that details an explicit acquisition policy. As an example, at $t=1$, the tree policy directs us to first inspect the IIP affiliation subscale acquired at onboarding and checks if the subscale score is either greater than 0.12. If so, we will then consult the AUDIT alcohol screening total score. If the value is greater than 16.50, we then move down the tree and check the person's YAACQ control subscale score. Depending on the value of that score, we will then decide to acquire the variables in either P8 or P12. The other trees can be interpreted similarly. One key thing to note is that we have explicit branches in the tree that determines what should be done if a particular feature of interest is not acquired (i.e. `NA'). As a concrete example, let us consider the tree at $t = 3$. If we did not ask about non-drink plans at the first time step (because the person's demographics led us to choose P12 instead of P8), we would then follow the `NA' branch to then consider the sadness variable at the second timepoint. There are two different types of `NA' branches in Figure \ref{fig:cheears_tree}. The `NA' branches in brown are obtained from the recursive partitioning algorithm. The `NA' branches colored in green with faded leaves are those where no training examples ever reached the state, so we assign the majority panel at training to the 'NA' branch there. To account for this, we assign the panel with the largest proportion between the `$\leq$' and `$>$' branches. Clearly, such a tree policy provides an explicit, interpretable policy that is readily executable.

\begin{figure}[h!]
    \centering
    \includegraphics[width=1\linewidth]{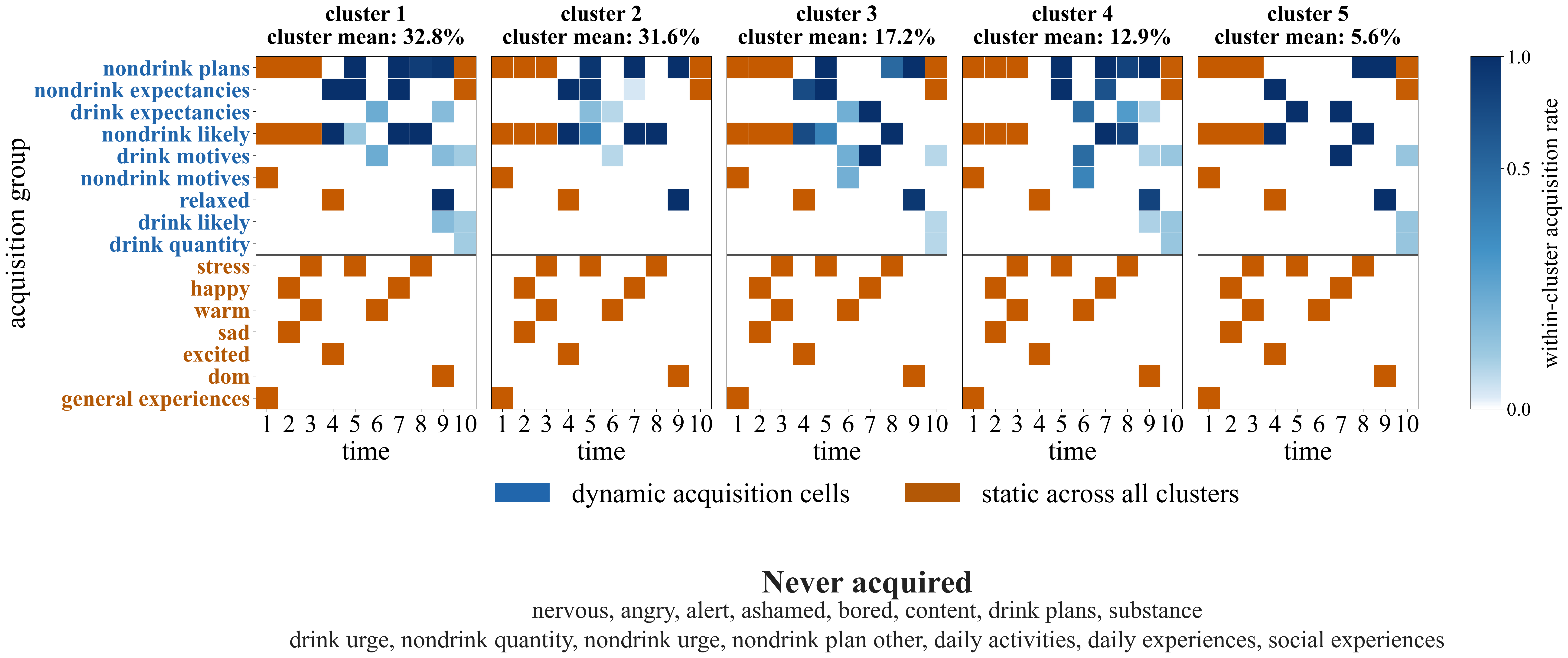}
    \caption{Heatmap of representative acquisition patterns on CHEEARS test set. We cluster on acquisition trajectories and plotted the centroids. Color shades indicate acquisition rate. Blue regions are where cross-cluster trajectories differs and orange cells are where the cell is always acuired for every participant across the clusters.}
    
\label{fig:kmeans_centroids}
\end{figure}

To visualize the acquisition trajectories, K-means was applied to each test participant’s complete binary acquisition trajectory. For each person, we flattened the $10\times31$ time-by-feature-group matrix into one vector and fit $K=5$ Euclidean K-means clusters. The displayed centroid for each cluster is therefore the mean acquisition rate, not an individual participant. We order feature groups on the y axis by cross-cluster trajectory variance. Orange cells indicate acquisition rates of at least .95 in every cluster and these same feature-timesteps are acquired in every cluster; blue cells indicate cluster-specific acquisition patterns. We did not plot features that are never acquired and listed them below the heatmap.

The acquisition patterns are shown in Figure \ref{fig:kmeans_centroids}. The five clusters contain 32.8\%, 31.6\%, 17.2\%, 12.9\%, and 5.6\% of test participants. Their mean acquisition costs are very similar (42.46–42.86), indicating that the clusters differ primarily in which information is acquired and when, rather than in total cost used. The results show a common acquisition in some features alongside meaningful personalization. The variables measuring affect and general experiences  form a nearly deterministic shared schedule across all clusters, shown by the orange cells. In contrast, nondrinking plans, expectancies, likelihood, motives, and the rarer drinking-likelihood/quantity variable groups vary across clusters and decision times. This shows TREACT tailors the longitudinal questions to each participants.

% {\color{red}@Bing, could you fill in qualitative results discussion on the tree? }
% brown n/a branches are from fitting on training data and purple n/a branches are when no training data are in this state so we assign the majority panel to it 

% {\color{red} @Yunni: Insert model distillation results here}

\section{Limitations}
Despite its promises, there are several key limitations in our proposed approach. The first is that there is some inherent ambiguity as to how the costs $\bm{c}_s$ and $\bm{c}_x$ are specified in the model. In the current setup, we have defined the cost of each variable in an abstract manner to capture any combination of effort, time or risk to the participant when responding to said item. This can be difficult to specify in practice. For example, some EMA studies embed entire cognitive tasks \parencite{sliwinski2018reliability} as part of the daily ambulatory assessments, and it can be challenging to quantify the relative costs of such tasks to measures of daily affect whose stimuli usually consists of a single target word \parencite{haney2023measuring}. Moreover, such costs can also vary by participant. As an example, items assessing trauma-related experiences can be more emotionally taxing on individuals with a history of trauma. The current setup of having a single person-invariant numeric cost can be challenging to specify and defend, although recent advances where such costs are learned from observed behavior may help alleviate this concern \parencite{lee2024bridging, wulfmeier2017large}. 

Another prominent limitation of our method is that the we have completely sidestepped typical psychometric considerations (e.g. reliability, factorial structure, measurement precision) when selecting items to present in an EMA study. Many instruments used in EMA studies are often developed and validated based on such psychometric principles, and choosing variables purely based on their predictive power can seem almost antithetical to conventional psychometric wisdom. This is particularly acute if the EMA data collected is used for secondary data analysis where the criterion variable is not of interest. These concerns can be alleviated to some extent in our current framework. One straightforward option is to generate masks where entire blocks of items belonging to the same battery are selected together. By selecting entire sets of variables, the psychometric properties of such scales can be preserved. An alternative would be to augment the loss function we have with a reconstruction loss \parencite{wang2022comprehensive} such that imputations based on the acquired variables can faithfully reconstruct the data as if everything was collected \parencite{yoon2018gain}. A third possibility is to explicitly integrate IRT-based methods designed for EMA studies \parencite{schneider2024just} with our prediction-based method. Indeed, this would be in line with recent work combining machine learning with IRT in the cross-sectional setting \parencite{gonzalez2025combining}. Other works have also demonstrated how confirmatory IRT models can be estimated neural networks \parencite{molenaar2025autoencoders}, thus opening up the possibility of integrating both prediction-focused and measurement-focused methods in a common neural framework. We leave such explorations for future research. 

A key limitation to both REACT and TREACT is that we relied on a rolling window procedure to train these models. This reflects a constraint of the REACT model architecture that we distill from. REACT takes flattens the temporal features into a fixed-length input, and cannot handle inputs of varying length. Naturally, TREACT also inherits this constraint. In this sense, the evaluations that we have conducted so far are based on this windowing preprocessing procedure. The resulting policies displayed in our figures are also validated only on the same horizon length. More research is needed to generalize our setup to time horizons with varying lengths. One promising approach to resolving this fixed window-length issue is to use Recurrent Neural Networks in the REACT architecture. That being said, we believe that the interpretable policies produced by TREACT are still useful for EMA studies with more modest time lengths (e.g. 7 days or 2 weeks).

Our tree-based model distillation procedure also comes with limitations. Like other interpretable methods, although it buys us interpretability, this comes at a cost of predictive power, but this is well known to be the cost of interpretability \parencite{Bertsimas2019ThePO}. In the empirical example, we see that there is a gap between the predictive power of TREACT and REACT. This can be ameliorated to some extent by using trees with more depth. However, increasing the tree depth would also make the trees harder to read. In practice, the choice of tree depth would depend on the particular application at hand. If the goal is to maximise predictive power while maintaining interpretability, a larger tree depth would be warranted. On the other hand, other applications might prefer to distill a tree policy so that the acquisition process can be further refined by domain experts. Such applications might instead favour shallower trees; a shallower tree policy is easier to interpret and is more amenable for communication purposes. Future work can further study how tree depth affects variable selection in realistic EMA settings.

Notwithstanding these limitations, TREACT provides a useful foundation for interpretable, time-varying feature acquisition in complex psychometric data. Future work addressing the limitations discussed above may further improve its flexibility, robustness, and applicability to EMA studies.

% Our experimental setup uses fixed 10-day windows. This reflects a constraint of the REACT architecture we distil, whose predictor takes a flattened fixed-length input; attention- or RNN-based acquisition models would remove it. TREACT inherits the constraint through its feature encoding rather than through the tree fitting: split columns are indexed by absolute visit, col(f, s) = f·T + s, and the protocol comprises one tree per time point, so a fitted protocol is defined only for the horizon it was trained on. The sequential fitting procedure and the panel-CE splitting criterion are themselves horizon-agnostic; extending TREACT to variable-length studies would require re-expressing splits over relative history, most-recent values, or time-since-last-acquisition — which would also make the printed rules independent of calendar position. We have not evaluated this variant.

\section{Conclusion}

In this paper, we introduced TREACT, a method for dynamic, personalized, and interpretable test-time variable selection in EMA studies. EMA data are inherently dynamic, and the variables most useful for prediction may differ across participants and measurement occasions. Collecting every variable at every occasion can therefore impose unnecessary burden on participants. TREACT addresses this challenge by distilling a neural-network-based longitudinal active feature acquisition policy into a sequence of interpretable decision trees. The resulting trees provide concrete, person-specific rules for determining which variables to acquire over time while balancing predictive accuracy against participant burden.

Our numerical studies demonstrated several benefits of this approach. In the simulation study, TREACT reliably recovered dynamically relevant sets of variables, whereas static acquisition regimes that could not adapt across participants performed less effectively. In the empirical application to alcohol consumption, TREACT retained favorable predictive accuracy while reducing acquisition costs through selective feature acquisition, skipped measurement occasions, and early termination of acquisition. These findings also demonstrate that an opaque neural acquisition policy can be approximated by an interpretable tree-based protocol without sacrificing most of its predictive and cost-saving benefits.

Taken together, our findings suggest that TREACT can produce coherent, interpretable, and cost-efficient acquisition policies for complex EMA studies. More broadly, we hope this work encourages the development of variable-selection methods that account for both the temporal structure of intensive longitudinal data and the practical demands placed on participants.

% We directly address this by learning time-varying acquisition masks that are personalized to each individual while reducing participant burden. 

% The acquisition policies that generate these masks can also be applied to new participants to select informative subsets of variables while reducing participant burden. In sum, we hope that our work will 

% Discussion 
%% Discuss choice of costs --- right now costs are naive; EMA might consist of not just self-report, but also cognitive tasks that might take more time / effort per question; effort might also differ per person. 
%% Discuss possibility of merging with CAT traditions, where optimizing for prediction might benefit from some optimization for measurement, cite some paeprs, and mention possibility 
%% We need access to a large-ish dataset 

\printbibliography

\end{document}